\documentclass[conference]{IEEEtran}
\IEEEoverridecommandlockouts
\usepackage{mathtools}
\usepackage{cite}
\usepackage{amsmath,amssymb,amsfonts}
\usepackage{algorithmic}
\usepackage{graphicx}
\usepackage{textcomp}
\usepackage{xcolor}
\usepackage{multirow}
\usepackage{caption}
\usepackage{subcaption}
\def\BibTeX{{\rm B\kern-.05em{\sc i\kern-.025em b}\kern-.08em
    T\kern-.1667em\lower.7ex\hbox{E}\kern-.125emX}}

\begin{document}

\title{Design of Turing Systems with Physics-Informed Neural Networks\\
}

\author{\IEEEauthorblockN{Jordon Kho}
\IEEEauthorblockA{\textit{School of Computer Science and Engineering} \\
\textit{Nanyang Technological University}\\
Singapore, Singapore \\
jkho006@e.ntu.edu.sg}
\and
\IEEEauthorblockN{Winston Koh}
\IEEEauthorblockA{\textit{Institute of Bioengineering and Bioimaging, A*STAR } \\
\textit{Bioinformatics Institute, A*STAR}\\
Singapore, Singapore \\
Winston\_koh@ibb.a-star.edu.sg}
\and
\IEEEauthorblockN{Jian Cheng Wong}
\IEEEauthorblockA{\textit{Dept of Fluid Dynamics} \\
\textit{Institute of High Performance Computing}\\
Singapore, Singapore \\
wongj@ihpc.a-star.edu.sg}
\and
\IEEEauthorblockN{Pao-Hsiung Chiu}
\IEEEauthorblockA{\textit{Dept of Fluid Dynamics} \\
\textit{Institute of High Performance Computing}\\
Singapore, Singapore \\
chiuph@ihpc.a-star.edu.sg}
\and
\IEEEauthorblockN{Chin Chun Ooi}
\IEEEauthorblockA{\textit{Dept of Fluid Dynamics} \\
\textit{Institute of High Performance Computing}\\
\textit{Centre for Frontier AI Research}\\
Singapore, Singapore \\
ooicc@ihpc.a-star.edu.sg}
}

\maketitle

\begin{abstract}
Reaction-diffusion (Turing) systems are fundamental to the formation of spatial patterns in nature and engineering. These systems are governed by a set of non-linear partial differential equations containing parameters that determine the rate of constituent diffusion and reaction. Critically, these parameters, such as diffusion coefficient, heavily influence the mode and type of the final pattern, and quantitative characterization and knowledge of these parameters can aid in bio-mimetic design or understanding of real-world systems. However, the use of numerical methods to infer these parameters can be difficult and computationally expensive. Typically, adjoint solvers may be used, but they are frequently unstable for very non-linear systems. Alternatively, massive amounts of iterative forward simulations are used to find the best match, but this is extremely effortful. Recently, physics-informed neural networks have been proposed as a means for data-driven discovery of partial differential equations, and have seen success in various applications. Thus, we investigate the use of physics-informed neural networks as a tool to infer key parameters in reaction-diffusion systems in the steady-state for scientific discovery or design. Our proof-of-concept results show that the method is able to infer parameters for different pattern modes and types with errors of less than 10\%. In addition, the stochastic nature of this method can be exploited to provide multiple parameter alternatives to the desired pattern, highlighting the versatility of this method for bio-mimetic design. This work thus demonstrates the utility of physics-informed neural networks for inverse parameter inference of reaction-diffusion systems to enhance scientific discovery and design.
\end{abstract}

\begin{IEEEkeywords}
Data-driven discovery, non-linear partial differential equations, parameter inference, physics-informed neural networks, reaction-diffusion
\end{IEEEkeywords}

\section{Introduction}

Reaction-diffusion (RD) models, commonly referred to as Turing models, have been widely studied in the realms of chemistry, biology and engineering and are hypothesised to be representative of the processes involved in the formation and evolution of many naturally-occurring spatial patterns. Turing first proposed that a system with two chemicals, such as an activator and inhibitor, could generate spatial patterns due to the reaction and diffusion between these chemicals from an initial near-homogeneous state in his seminal work on Turing models \cite{turing1990chemical}. Since then, numerous skin patterns found in nature, including the marine angelfish \cite{kondo1995,kondo2010}, zebrafish \cite{asai1999} and cats \cite{bard1981}, have been found to evolve in accordance with this mathematical framework. Similarly, RD processes have been critical to the synthesis of nanostructures \cite{yin2004} and various forms of self-assembly and self-organisation \cite{boekhoven2015,mann2009}. 

These RD systems are described by a system of equations comprising partial differential equations (PDEs) - one for each constituent chemical in the system - which fundamentally describe individual reaction and diffusion processes. As the parameters in this system of equations heavily influence the pattern mode and type (e.g. stripes, spots), obtaining their values can be desirable to enhance our fundamental understanding and/or facilitate bio-memetic design across many biological and chemical applications. 

For example, Miyazawa, Okamoto and Kondo used a RD model for zebrafish skin patterns to demonstrate how the interaction across multiple parameters affect the final pattern obtained \cite{miyazawa2010}. They showed how variations in parameters such as the diffusion and reaction coefficients lead to the formation of different pattern types such as spots and labyrinthine stripes. Inverse inference of parameter values from observed biological patterns can then provide insight into the biological constituents that are at play. Similarly, the development of spatial patterns in many chemical systems follow similar mathematical models. For instance, the Lengyel-Epstein model used for modelling the chloride-iodide-malonic acid-starch (CIMA) reaction - one of the first reactions which presented experimental evidence of Turing patterns - takes in four input parameters \cite{maini1997,ni2005turing}. In order to generate desired patterns of specific modes and types, knowledge of the underlying parameter values such as diffusion coefficient are essential to the judicious choice of constituent chemicals (with appropriate diffusion and reaction characteristics) so as to avoid multiple rounds of experimental trial-and-error. However, these values are difficult to derive \textit{a priori}, especially in design. Critically, numerical methods to solve such inverse problems are computationally expensive and may be difficult to implement for complex systems due to the under-determined nature of such inverse problems \cite{kaltenbacher2020inverse}. 

In recent years, Physics-Informed Neural Networks (PINNs) have become an increasingly popular tool for physics-informed learning \cite{karniadakis2021physics,chiu2022can}. Fundamentally, PINNs use a neural network as a universal approximator for physical systems and incorporate a set of governing physical laws to regularize the learned model during training \cite{raissi2019}. In particular, PINNs have been shown to perform well for two main types of problems: i) data-driven solution and ii) data-driven discovery of PDEs. The former involves the use of a forward model to approximate solutions to a specific PDE, while the latter involves the use of an inverse model to infer PDE parameters which best describe the observed data. Among others, Raissi, Perdikaris and Karniadakis have demonstrated the ability of the inverse model to infer parameters with low errors in both two-dimensional Navier-Stokes equations and Korteweg-de Vries equation when given scattered and noisy data \cite{raissi2019}. Arthurs and King further investigated the use of PINNs to find model parameters in Navier-Stokes equations for specific physical properties \cite{arthurs2020}. In this instance, they wanted to design a tube shape to produce a desired change in pressure, and the PINN was able to give predictions with high accuracy. Separately, Eivazi, Tahani, Schlatter and Vinuesa employed PINNs to solve Reynolds-Averaged Navier-Stokes equations for turbulent flow across different boundary layers with small errors \cite{eivazi2021}. These results highlight the viability of PINNs as a method for model parameter inference across multiple domains such as flow and electromagnetics. 

Hence, in this work, we elect to study the ability of PINNs to solve the problem of data-driven discovery of PDEs in RD systems. Critically, RD systems have been relatively less studied in the PINN literature with only a few successful demonstrations of forward modelling \cite{giampaolo2022physics}. This could be related to observations in other prior work suggesting that non-linear systems such as RD processes may require fairly sophisticated methods for successful modelling \cite{wang2021eigenvector,krishnapriyan2021characterizing}. Interestingly, our results suggest that the data-driven inference of model parameters via PINNs is not unduly hampered by these failure modes and that Turing systems can indeed be successfully designed via this PINN framework.

\section{Materials and Methods}

The methodology is as follows: Firstly, a numerical model was developed to generate Turing patterns in two-dimensional space. These patterns were subsequently provided to the inverse PINN for inferring the system parameters undergirding the RD equations. Finally, the PINN-derived parameter values were validated with the numerical model to ensure that they did indeed generate the corresponding patterns, thus completing a theoretical design loop.

\subsection{Theoretical and Numerical Model}

The reaction-diffusion system being studied in this work is based on prior work by Barrio, Varea, Arag{\'o}n, and Maini \cite{barrio1999}. More specifically, this 2-component system is governed by the following equations:

\begin{equation}
\label{u-pde}
    \frac{\partial u}{\partial t}
    = D_{1} D_{2} \nabla^2 u
    + \alpha u \left(1 - r_{1} v^2 \right)
    + v \left(1 - r_{2} u \right)
\end{equation}
\begin{equation}
\label{v-pde}
    \frac{\partial v}{\partial t}
    = D_{2} \nabla^2 v
    + \beta v \left(1 + \frac{\alpha r_{1}}{\beta} u v \right)
    + u \left(\gamma + r_{2} v \right)
\end{equation}
where $u(t,x,y)$ and $v(t,x,y)$ are the concentrations of chemicals $u$ and $v$ respectively, $D_{1}$ and $D_{2}$ are parameters related to the individual diffusion coefficients, $\alpha, \beta$ and $\gamma$ are system parameters, and $r_{1}$ and $r_{2}$ are interaction parameters which represent cubic and quadratic terms respectively. It is also given that $\gamma = -\alpha$.

The Turing patterns studied in this work were generated on a $50 \times 50$ grid according to the governing equations described in \eqref{u-pde} and \eqref{v-pde}. A $2^{nd}$ order central difference scheme and $1^{st}$ order explicit Euler scheme was used for discretization of the spatial and time derivatives and evaluation of the numerical model.

Two patterns - one with stripes and one with spots - were recreated from the set of parameters published in Barrio et al. to validate our numerical model. Both patterns chosen have a dominant mode, $k = 0.420$, as defined by the parameter set: $D_{1} = 0.516, D_{2} = 2.00, \alpha = 0.899$ and $\beta = -0.910$. As reported by Barrio et al., interaction parameter values of $r_{1} = 3.500$ and $r_{2} = 0.000$ will generate a striped pattern while interaction parameter values of $r_{1} = 0.020$ and $r_{2} = 0.200$ will generate a spotted pattern.

Separately, a third parameter set comprising $D_{1} = 0.300, D_{2} = 2.000, \alpha = 0.700, \beta = -0.750, r_{1} = 3.500$ and $r_{2} = 0.000$ was defined to generate a striped pattern with a different dominant mode, $k = 0.600$, for additional evaluation with the PINN model.

For this work, only the steady-state result are studied. Hence, all numerical models were run till the time derivatives approached zero such that the asymptotic steady-state patterns could be extracted for each parameter set.

\subsection{Physics-Informed Neural Network}

The forward PINN works on partial differential equations of the form:

\begin{equation}
    N[h] = 0
\end{equation}
where $h(x,y)$ is the hidden solution and $N[\cdot]$ is any linear and non-linear combination of various spatial differential operators. While time derivatives can be included, time derivatives are excluded for this work which focuses on steady-state analysis. $N[\cdot]$ can further be parameterised such that the PINN now solves the problem of data-driven discovery of partial differential equations. The differential operator then takes the following form:

\begin{equation}
    N[h;\lambda] = 0
\end{equation}
where $h(x,y)$ is the hidden solution, $N[\cdot;\lambda]$ is the combination of all possible differential operators and $\lambda$ is the parameter set to be inferred. A neural network is used to approximate $h(x,y)$ which results in an inverse PINN of the form:

\begin{equation}
    f := N[h;\lambda]
\end{equation}

By substituting $N[\cdot;\lambda]$ with \eqref{u-pde} and \eqref{v-pde}, $f_{u}$ and $f_{v}$ are defined as:

\begin{equation}
\label{u-pinn}
    f_{u} = D_{1} D_{2} \nabla^2 u
    + \alpha u \left(1 - r_{1} v^2 \right)
    + v \left(1 - r_{2} u \right)
\end{equation}
\begin{equation}
 \label{v-pinn}
    f_{v} = D_{2} \nabla^2 v
    + \beta v \left(1 + \frac{\alpha r_{1}}{\beta} u v \right)
    + u \left(\gamma + r_{2} v \right)
\end{equation}

The overall loss function for neural network training is then the sum of mean squared errors (MSEs) of three components:

\begin{equation}
\label{total-mse}
    MSE = MSE_{h} + w_{f} MSE_{f} + MSE_{bc}
\end{equation}
where $MSE_{h}$, $MSE_{f}$ and $MSE_{bc}$ are the MSEs of $h(x,y)$, $f(x,y)$ and the boundary data points respectively, and $w_{f}$ is the relative weight of $MSE_{f}$. 

$MSE_{h}$ is defined as:

\begin{equation}
\label{data-mse}
    MSE_{h} = \frac{1}{N_{h}} \sum_{i=1}^{N_{h}} \left(h^i_{pred} - h^i_{data} \right)^2
\end{equation}
where $N_{h}$ is the number of training data points, $\{h^{i}_{pred}\}_{i=1}^{N_{h}}$ are the predicted $h$ values, and $\{h_{i}\}_{i=1}^{N_{h}}$ are the actual $h$ values.

$MSE_{f}$ is defined as:
    
\begin{equation}
\label{pde-mse}
    MSE_{f} = \frac{1}{N_{f}} \sum_{i=1}^{N_{f}} \left(f \left(x_{f}^i,y_{f}^i \right) \right)^2
\end{equation}
where $N_{f}$ is the number of sampled points within the grid and $\{f(x_{f}^i,y_{f}^i)\}_{i=1}^{N_{f}}$ are the $f$ values at select collocation points. 

$MSE_{bc}$ is defined as:

\begin{equation}
\label{bc-mse}
    MSE_{bc} = \frac{1}{N_{bc}} \sum_{i=1}^{N_{bc}} \left(f \left(x_{bc}^i,y_{bc}^i \right) \right)^2
\end{equation}
where $N_{bc}$ is the number of sampled points at the boundaries. 

The domain of interest extends for 200 dimensionless units in both x and y directions, spanning $-100$ to $100$. After hyperparameter tuning, the architecture of the final inverse PINN model used for all experiments in this work is as follows: number of hidden layers $= 4$, number of neurons per hidden layer $= 64$, learning rate $= 2.50 \times 10^{-4}$, batch size $= 25$, $N_{f} = 2500$, $N_{bc} = 200$ and $w_{f} = 10$. The activation function used in each neuron is the hyperbolic tangent function $tanh$.

\subsection{Description of Parametric Cases}

Two groups of experiments were conducted with each group consisting of different sets of parameters as shown in Table \ref{tab: param_sets}. 

Group 1 consists of two sets - A and B - and served as a baseline for evaluating the PINN's performance in inferring RD parameters. Each set is restricted to just two related parameters - $D_{1}, D_{2}$ and $\alpha, \beta$. 

Group 2 consists of three sets - C, D and E - and was designed to simulate different situations where only certain parameters are known. For example, Sets C and D are useful when $r_{1}$ is known. Out of the six parameters, $D_{1}, D_{2}, \alpha$ and $\beta$ were chosen for inference because they are more important in determining the pattern mode, whereas $r_{1}$ and $r_{2}$ can generally be inferred from the type of pattern  observed or desired (e.g. stripes or spots) and may not be always needed. Nonetheless, $r_{1}$ was included in Set E to simulate possible scenarios where it is not known and more refined values need to be inferred. 

In addition, our preliminary results from Set C suggest that models involving \eqref{u-pinn} and \eqref{v-pinn} alone are under-constrained and have a large solution space. This is also indicative of the issues confronted by numerical methods in inverting a many-to-one complex mapping. Hence, one parameter was fixed to constrain the solution space for better comparison. $D_{2}$ was chosen since it is indicative of the scale of the spatial pattern, and hence, is a parameter which is more readily obtainable from any real-world system or observation \cite{barrio1999}. 

For all sets, the parameters were initialised to $0.000$ with the exception of $\beta$, which was set to $1.000$ since it is a denominator as per \eqref{v-pde}.

\begin{table}[htbp]
\caption{Parameter Sets}
\begin{center}
\begin{tabular}{|c|c|c|c|c|c|c|}
\hline
\multirow{2}{*}{\textbf{Group}} & \multirow{2}{*}{\textbf{Set}} & \multicolumn{5}{|c|}{\textbf{Parameter Set, $\lambda$}} \\
\cline{3-7}
{} & {} & $D_{1}$ & $D_{2}$ & $\alpha$ & $\beta$ & $r_{1}$ \\
\hline
\multirow{2}{*}{1} & A & \checkmark & \checkmark & {} & {} & {} \\
\cline{2-7}
{} & B & {} & {} & \checkmark & \checkmark & {} \\
\hline
\multirow{3}{*}{2} & C & \checkmark & \checkmark & \checkmark & \checkmark & {} \\
\cline{2-7}
{} & D & \checkmark & {} & \checkmark & \checkmark & {} \\
\cline{2-7}
{} & E & \checkmark & {} & \checkmark & \checkmark & \checkmark \\
\hline
\end{tabular}
\label{tab: param_sets}
\end{center}
\end{table}

\section{Results and Discussions}

\subsection{Turing Patterns}

\begin{figure}
    \centering
    \begin{subfigure}{0.3\textwidth}
        \centering
        \includegraphics[width=\textwidth]{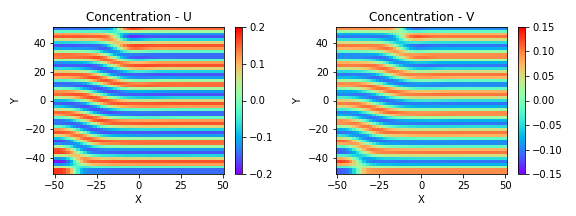}
        \caption{Striped pattern of mode k = 0.420 (Pattern P)}
        \label{fig:Test Case A}
    \end{subfigure}
    \hfill
    \begin{subfigure}{0.3\textwidth}
        \centering
        \includegraphics[width=\textwidth]{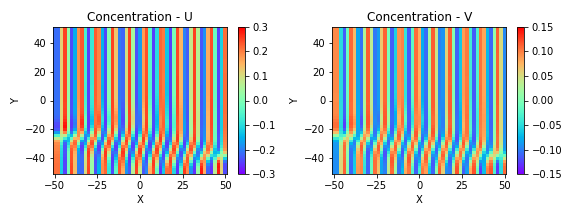}
        \caption{Striped pattern of mode k = 0.600 (Pattern Q)}
        \label{fig:Test Case E}
    \end{subfigure}
    \begin{subfigure}{0.3\textwidth}
        \centering
        \includegraphics[width=\textwidth]{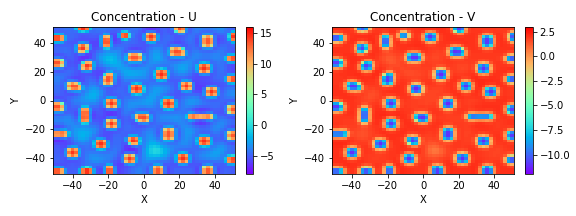}
        \caption{Spotted pattern of mode k = 0.420 (Pattern R)}
        \label{fig:Test Case Low}
    \end{subfigure}
        \caption{Concentrations of $u$ and $v$ in the steady-state}
        \label{fig:Turing Patterns}
\end{figure}

Three Turing patterns (P, Q and R) were generated using the numerical model as described above, and their initial parameters are shown in Table \ref{tab: ini_params}. It is known that stripes are favoured by the cubic term $r_{1}$ while spots are favoured by the quadratic term $r_{2}$ \cite{ermentrout1991}. Hence, Patterns P and Q are striped since their $r_{1}$ values are greater than their $r_{2}$ values, and Pattern R is spotted since its $r_{1}$ value is less than its $r_{2}$ value. Fig. \ref{fig:Test Case A}, Fig. \ref{fig:Test Case Low} and Fig. \ref{fig:Test Case E} show the steady-state concentrations of $u$ and $v$ of Patterns P, Q and R respectively.

\begin{table}[htbp]
\caption{Initial Parameters of Turing Patterns}
\begin{center}
\begin{tabular}{|c|c|c|c|c|c|c|c|}
\hline
\multirow{2}{*}{\textbf{Pattern}} & \multirow{2}{*}{\textbf{Mode}} & \multicolumn{6}{|c|}{\textbf{Initial Parameters}} \\
\cline{3-8} 
{} & {} & $D_{1}$ & $D_{2}$ & $\alpha$ & $\beta$ & $r_{1}$ & $r_{2}$ \\
\hline
P & $0.42$ & $0.516$ & $2.0$ & $0.899$ & $-0.91$ & $3.50$ & $0.0$ \\
\hline
Q & $0.60$ & $0.300$ & $2.0$ & $0.700$ & $-0.75$ & $3.50$ & $0.0$ \\
\hline
R & $0.42$ & $0.516$ & $2.0$ & $0.899$ & $-0.91$ & $0.02$ & $0.2$ \\
\hline
\end{tabular}
\label{tab: ini_params}
\end{center}
\end{table}

In addition to reporting the inferred parameter sets, data loss will also be reported in the subsequent sections as a measure for assessing the PINN's performance in accurately reproducing the provided pattern. 

\subsection{Baseline Performance}

Since Group 1 was intended as a baseline for the PINN's inference performance, only one pattern (Pattern P) was used for evaluation in this first study. 

For Set A, the data loss is $7.3 \times 10^{-6}$ and the inferred values are $0.511$ with an error of $1.1\%$ for $D_{1}$, and $1.874$ with an error of $6.3\%$ for $D_{2}$. The PINN had even lower inference errors with Set B, where $\alpha$ was found to be $0.911$ with an error of $1.4\%$, and $\beta$ was found to be $-0.915$ with an error of $0.5\%$. The data loss is slightly higher than that of Set A at $8.4 \times 10^{-6}$. 

This suggests that the inverse inference for $D_{1}$ and $D_{2}$ tends to be a slightly more difficult problem for the PINN model, relative to $\alpha$ and $\beta$. This could also suggest that the patterns observed are particularly sensitive to the values of $D_{1}$ and $D_{2}$, as these values need to be more accurately inferred in order for the loss to decrease.

\subsection{Parameter Inference with Pattern P}

For Group 2 experiments, each set was independently initialized and run eight times and the mean and variance of the inferred parameters across the eight runs were recorded.

The mean and variance of the inferred Set C parameters of Pattern P are shown in Table \ref{tab: set_C_P}. For all runs, the PINN was unable to accurately infer the exact same initial set of parameters and has mean errors of $\approx 88\%$. Yet, its mean data loss is $3.7 \times 10^{-6}$ which is lower than that of the baselines ($7.3 \times 10^{-6}$ and $8.4 \times 10^{-6}$). 

This suggests that the PINN inferred another valid set of parameters which can produce the same pattern. The small relative variances in the inferred parameters also indicate that the obtained solutions across the 8 runs are close to one another, and that the optimization process has converged successfully, despite the relatively large difference in parameters. This further shows that \eqref{u-pinn} and \eqref{v-pinn} define a system with many-to-one mapping, where multiple parameter sets are able to generate any particular pattern mode and type in the steady-state.

\begin{table}[htbp]
\caption{Inferred Set C Parameters of Pattern P}
\begin{center}
\begin{tabular}{|c|c|c|c|c|}
\hline
\multirow{2}{*}{\textbf{Measure}} & \multicolumn{4}{|c|}{\textbf{Inferred Parameters}} \\
\cline{2-5}
{} & {$D_{1}$} & {$D_{2}$} & {$\alpha$} & {$\beta$} \\
\hline
{Mean} & {$0.645$} & {$0.242$} & {$0.715$} & {$-0.999$} \\
\hline
{Variance} & {$1.3 \times 10^{-4}$} & {$2.0 \times 10^{-3}$} & {$4.3 \times 10^{-5}$} & {$7.0 \times 10^{-7}$} \\
\hline
{Error (\%)} & {$25.0$} & {$87.9$} & {$20.4$} & {$9.8$} \\
\hline
{Data Loss} & \multicolumn{4}{|c|}{$3.7 \times 10^{-6}$} \\
\hline
\end{tabular}
\label{tab: set_C_P}
\end{center}
\end{table}

\begin{figure}
    \centering
    \begin{subfigure}{0.5\textwidth}
        \centering
        \includegraphics[width=\textwidth]{TestCaseA.png}
        \caption{Original Pattern P with Set C parameters $D_{1} = 0.516, D_{2} = 2.000, \alpha = 0.899, \beta = -0.910$}
        \label{fig:Original A d1d2ab}
    \end{subfigure}
    \hfill
    \begin{subfigure}{0.5\textwidth}
        \centering
        \includegraphics[width=\textwidth]{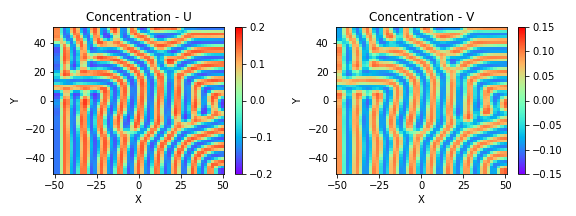}
        \caption{Alternative Pattern P with Set C parameters $D_{1} = 0.618, D_{2} = 0.188, \alpha = 0.707, \beta = -0.997$}
        \label{fig:Alternative A d1d2ab}
    \end{subfigure}
        \caption{Alternative solutions of Pattern P with different $D_{1}, D_{2}, \alpha, \beta$ values}
        \label{fig:A d1d2ab}
\end{figure}

To validate this hypothesis, a set of parameters obtained from one of the runs was provided to the numerical model and an alternative solution was generated. Fig. \ref{fig:Original A d1d2ab} shows the original pattern while Fig. \ref{fig:Alternative A d1d2ab} shows the alternative pattern. While the alternative pattern does not contain straight and distinct stripes like those in the original pattern, they appear almost identical in terms of stripe periodicity, thickness, and magnitude. The differences in L2 norm of the concentrations of $u$ and $v$ between the original and alternative patterns were found to be $0.2\%$ and $1.3\%$ respectively. This highlights the PINN's ability to provide alternative solutions with $\approx 1\%$ difference in magnitude. Critically, the variance in inferred parameters within this set remain low ($7.0 \times 10^{-7}$).

Importantly, it is known that even the same parameter set can evolve different final patterns depending on the initialization, although the final patterns will tend to maintain the same macro properties such as stripe periodicity and amplitude. Hence, it is unsurprising in this instance that the stripes' position, orientation and curvature are different due to the stochastic nature of the starting seed used for initialising the concentrations of $u$ and $v$. It is thus possible for different 'correct' parameter sets to be obtained using stochastic initialization, thus opening up the space for bio-memetic design. 

To simulate the cases where $D_{2}$ is specified, $D_{2}$ was fixed in Sets D and E. Fixing $D_{2}$ also has the effect of constraining the solution space further for this many-to-one problem. The mean and variance of the inferred Set D parameters of Pattern P are shown in Table \ref{tab: set_D_P}. Unlike Set C, the PINN was then able to accurately infer the parameters across all runs with errors from $0.6$ to $3.5\%$ and a data loss of $5.9 \times 10^{-6}$, further highlighting that the inference of specific parameter sets can be effectively obtained with a better constrained problem (e.g. when $D_{2}$ is known).

\begin{table}[htbp]
\caption{Inferred Set D Parameters of Pattern P}
\begin{center}
\begin{tabular}{|c|c|c|c|}
\hline
\multirow{2}{*}{\textbf{Measure}} & \multicolumn{3}{|c|}{\textbf{Inferred Parameters}} \\
\cline{2-4}
{} & {$D_{1}$} & {$\alpha$} & {$\beta$} \\
\hline
{Mean} & {$0.498$} & {$0.904$} & {$-0.896$} \\
\hline
{Variance} & {$4.5 \times 10^{-4}$} & {$9.0 \times 10^{-5}$} & {$2.5 \times 10^{-4}$} \\
\hline
{Error (\%)} & {$3.5$} & {$0.6$} & {$1.5$} \\
\hline
{Data Loss} & \multicolumn{3}{|c|}{$5.9 \times 10^{-6}$} \\
\hline
\end{tabular}
\label{tab: set_D_P}
\end{center}
\end{table}

The mean and variance of the inferred Set E parameters of Pattern P are shown in Table \ref{tab: set_E_P}. Although the difference between the initial parameter set and the inferred parameter set is high at $42.0\%$, the mean data loss is low ($5.7 \times 10^{-6}$) and comparable to that of the baselines ($7.3 \times 10^{-6}$ and $8.4 \times 10^{-6}$). It is thus likely that the PINN has found other valid parametric solutions again.

\begin{table}[htbp]
\caption{Inferred Set E Parameters of Pattern P}
\begin{center}
\begin{tabular}{|c|c|c|c|c|}
\hline
\multirow{2}{*}{\textbf{Measure}} & \multicolumn{4}{|c|}{\textbf{Inferred Parameters}} \\
\cline{2-5}
{} & {$D_{1}$} & {$\alpha$} & {$\beta$} & {$r_{1}$} \\
\hline
{Mean} & {$0.494$} & {$0.914$} & {$-0.894$} & {$4.970$} \\
\hline
{Variance} & {$2.7 \times 10^{-4}$} & {$1.7 \times 10^{-4}$} & {$8.8 \times 10^{-5}$} & {$9.6 \times 10^{-1}$} \\
\hline
{Error (\%)} & {$4.2$} & {$1.6$} & {$1.8$} & {$42.0$} \\
\hline
{Data Loss} & \multicolumn{4}{|c|}{$5.7 \times 10^{-6}$} \\
\hline
\end{tabular}
\label{tab: set_E_P}
\end{center}
\end{table}

\begin{figure}
    \centering
    \begin{subfigure}{0.5\textwidth}
        \centering
        \includegraphics[width=\textwidth]{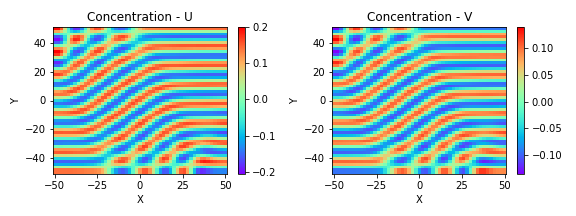}
        \caption{Original Pattern P with Set E parameters $D_{1} = 0.516, \alpha = 0.899, \beta = -0.910, r_{1} = 3.500$}
        \label{fig:Original A d1abr1}
    \end{subfigure}
    \hfill
    \begin{subfigure}{0.5\textwidth}
        \centering
        \includegraphics[width=\textwidth]{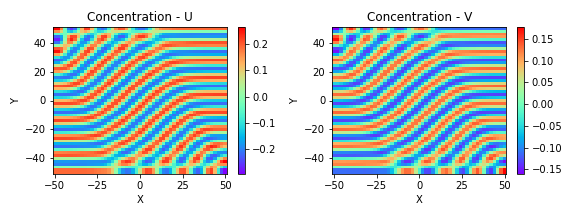}
        \caption{Alternative Pattern P with Set E parameters $D_{1} = 0.455, \alpha = 0.890, \beta = -0.872, r_{1} = 4.152$}
        \label{fig:Alternative A d1abr1}
    \end{subfigure}
        \caption{Alternative solutions of Pattern P with different $D_{1}, \alpha, \beta, r_{1}$ values}
        \label{fig:A d1abr1}
\end{figure}

Hence, as done above, the parameters obtained from one of the runs was provided to the numerical model to generate an alternative pattern. Fig. \ref{fig:Original A d1abr1} shows the original Pattern P with a different starting seed and Fig. \ref{fig:Alternative A d1abr1} shows the alternative solution. As mentioned, this also emphasizes the point that the same parameter set can generate seemingly different orientations under different initializations.

Similar to Set C, the stripe periodicity and thickness of both original and alternative patterns are alike. However, there was an increase in the L2 norms of the concentrations of $u$ and $v$, from $5.7$ to $8.0$ and from $3.8$ to $4.9$ respectively when $r_{1}$ increased from $3.500$ to $4.152$. From this result, it appears that $r_{1}$ controls the concentration scale where higher values lead to a greater difference in concentration peaks and troughs. Moreover, pattern appearance seems to be less sensitive to $r_{1}$ as the pattern retains its periodicity, thickness and type despite the slightly larger differences observed for $r_{1}$. On the other hand, $D_{1}, \alpha$ and $\beta$ have lower variances and seem to be particularly crucial in determining the pattern characteristics. These observations are also consistent with prior literature.

\subsection{Other Pattern Modes and Types}

From the results with Pattern P, it is evident that the PINN is capable of inferring RD parameters representing original and alternative solutions. Inferred parameters with low errors $< 10\%$ typically have lower errors when the stochastic optimization has led back to the original parameter set, whereas the inferred solutions with high errors and low data losses (with orders of magnitude of $10^{-6}$) typically belong to alternative parameter set solutions. 

Patterns Q and R were further tested with this methodology to verify that the PINN fares consistently well across other pattern modes and types.

First, the PINN was run with Set C parameters for Pattern Q and the mean and variance of the inferred parameters are shown in Table \ref{tab: set_C_Q}. Alternative solutions were presumably found since high inference errors of $\approx 99\%$ were obtained even though the data loss was small at $1.5 \times 10^{-5}$. This is a similar situation to that previously observed with Pattern P.

\begin{table}[htbp]
\caption{Inferred Set C Parameters of Pattern Q}
\begin{center}
\begin{tabular}{|c|c|c|c|c|}
\hline
\multirow{2}{*}{\textbf{Measure}} & \multicolumn{4}{|c|}{\textbf{Inferred Parameters}} \\
\cline{2-5}
{} & {$D_{1}$} & {$D_{2}$} & {$\alpha$} & {$\beta$} \\
\hline
{Mean} & {$0.410$} & {$0.026$} & {$0.468$} & {$-1.000$} \\
\hline
{Variance} & {$1.9 \times 10^{-3}$} & {$4.6 \times 10^{-4}$} & {$2.6 \times 10^{-5}$} & {$2.2 \times 10^{-7}$} \\
\hline
{Error (\%)} & {$36.7$} & {$98.7$} & {$33.1$} & {$33.3$} \\
\hline
{Data Loss} & \multicolumn{4}{|c|}{$1.5 \times 10^{-5}$} \\
\hline
\end{tabular}
\label{tab: set_C_Q}
\end{center}
\end{table}

\begin{figure}
    \centering
    \begin{subfigure}{0.5\textwidth}
        \centering
        \includegraphics[width=\textwidth]{TestCaseLow.png}
        \caption{Original Pattern Q with Set D parameters $D_{1} = 0.300, \alpha = 0.700, \beta = -0.750$}
        \label{fig:Original Low d1ab}
    \end{subfigure}
    \hfill
    \begin{subfigure}{0.5\textwidth}
        \centering
        \includegraphics[width=\textwidth]{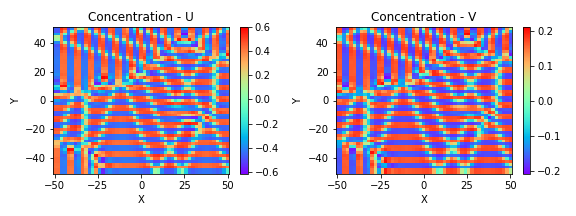}
        \caption{Alternative Pattern Q with Set D parameters $D_{1} = 0.218, \alpha = 0.662, \beta = -0.571$}
        \label{fig:Alternative Low d1ab}
    \end{subfigure}
        \caption{Alternative solutions of Pattern Q with different $D_{1}, \alpha, \beta$ values}
        \label{fig:Low d1ab}
\end{figure}

Next, Set D parameters of Pattern Q were inferred and the PINN was observed to provide both original solutions across most runs and an alternative solution for one run. The mean and variance of the inferred parameters for the original solution and the inferred values for the alternative solution are shown in Table \ref{tab: set_D_Q}. For parameters belonging to the original pattern, they result in lower inference errors from $2.4$ to $18.7\%$, while those belonging to the alternative pattern have higher errors from $5.5$ to $27.3\%$. The original and alternative solutions of Pattern Q are shown in Fig. \ref{fig:Original Low d1ab} and Fig. \ref{fig:Alternative Low d1ab} respectively. Again, we note that macro pattern descriptors such as the periodicity of the stripes are identical across the original and alternative parameter sets.

\begin{table}[htbp]
\caption{Inferred Set D Parameters of Pattern Q}
\begin{center}
\begin{tabular}{|c|c|c|c|c|}
\hline
\multirow{2}{*}{\textbf{Solution}} & \multirow{2}{*}{\textbf{Measure}} & \multicolumn{3}{|c|}{\textbf{Inferred Parameters}} \\
\cline{3-5}
{} & {} & {$D_{1}$} & {$\alpha$} & {$\beta$} \\
\hline
\multirow{4}{*}{Original} & {Mean} & {$0.244$} & {$0.683$} & {$-0.610$} \\
\cline{2-5}
{} & {Variance} & {$5.8 \times 10^{-5}$} & {$3.6 \times 10^{-5}$} & {$2.7 \times 10^{-4}$} \\
\cline{2-5}
{} & {Error (\%)} & {$18.7$} & {$2.4$} & {$18.6$} \\
\cline{2-5}
{} & {Data Loss} & \multicolumn{3}{|c|}{$2.1 \times 10^{-5}$} \\
\hline
\multirow{3}{*}{Alternate} & {Value} & {$0.218$} & {$0.662$} & {$-0.571$} \\
\cline{2-5}
{} & {Error (\%)} & {$27.3$} & {$5.5$} & {$23.8$} \\
\cline{2-5}
{} & {Data Loss} & \multicolumn{3}{|c|}{$4.0 \times 10^{-5}$} \\
\hline
\end{tabular}
\label{tab: set_D_Q}
\end{center}
\end{table}

\begin{figure}
    \centering
    \begin{subfigure}{0.5\textwidth}
        \centering
        \includegraphics[width=\textwidth]{TestCaseLow.png}
        \caption{Original Pattern Q with Set E parameters $D_{1} = 0.300, \alpha = 0.700, \beta = -0.750, r_{1} = 3.500$}
        \label{fig:Original Low d1abr1}
    \end{subfigure}
    \hfill
    \begin{subfigure}{0.5\textwidth}
        \centering
        \includegraphics[width=\textwidth]{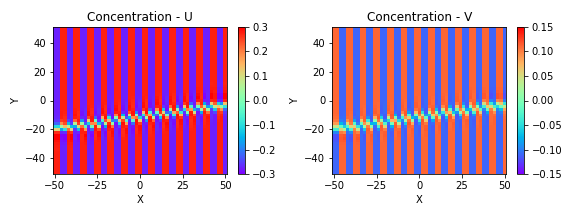}
        \caption{Alternative Pattern Q with Set E parameters $D_{1} = 0.252, \alpha = 0.708, \beta = -0.626, r_{1} = 5.982$}
        \label{fig:Alternative Low d1abr1}
    \end{subfigure}
        \caption{Alternative solutions of Pattern Q with different $D_{1}, \alpha, \beta, r_{1}$ values}
        \label{fig:Low d1abr1}
\end{figure}

The mean and variance for Set E parameters of Pattern Q are shown in Table \ref{tab: set_E_Q}. As $r_{1}$ has increased from $3.500$ to $5.982$, the L2 norms of the concentrations of $u$ and $v$ also increased from $9.7$ to $14.1$ and from $4.3$ to $5.7$ respectively. Once again, the scaling effect of $r_{1}$ is observed. The original and alternative solutions of Pattern Q are shown in Fig. \ref{fig:Original Low d1abr1} and Fig. \ref{fig:Alternative Low d1abr1} respectively. While the amplitudes are different, due to the changed values of $r_{1}$, we note that the periodicity remains consistent, as is expected from the relatively lower losses.

\begin{table}[htbp]
\caption{Inferred Set E Parameters of Pattern Q}
\begin{center}
\begin{tabular}{|c|c|c|c|c|}
\hline
\multirow{2}{*}{\textbf{Measure}} & \multicolumn{4}{|c|}{\textbf{Inferred Parameters}} \\
\cline{2-5}
{} & {$D_{1}$} & {$\alpha$} & {$\beta$} & {$r_{1}$} \\
\hline
{Mean} & {$0.261$} & {$0.716$} & {$-0.645$} & {$6.071$} \\
\hline
{Variance} & {$1.2 \times 10^{-4}$} & {$3.3 \times 10^{-4}$} & {$4.9 \times 10^{-4}$} & {$1.2 \times 10^{0}$} \\
\hline
{Error (\%)} & {$13.0$} & {$2.3$} & {$14.0$} & {$73.5$} \\
\hline
{Data Loss} & \multicolumn{4}{|c|}{$3.2 \times 10^{-5}$} \\
\hline
\end{tabular}
\label{tab: set_E_Q}
\end{center}
\end{table}

The final set of inferred parameters is Set D of Pattern R and the mean and variance of the inferred parameters are shown in Table \ref{tab: set_D_R}. No alternative solutions were returned as the parameters were inferred with high accuracy across all runs with errors from $2.2$ to $12.7\%$. Although the mean data loss is significantly greater at $2.0$, it can be attributed to the orders of magnitude higher concentrations of $u$ and $v$ for this system. The order of magnitude of the concentrations of $u$ and $v$ for Patterns P and Q is $\approx 10^{-1}$ while that for Pattern R is $\approx 10^{1}$. This 2 order difference in magnitude corresponds to the difference in mean data loss as well.

\begin{table}[htbp]
\caption{Inferred Set D Parameters of Pattern R}
\begin{center}
\begin{tabular}{|c|c|c|c|}
\hline
\multirow{2}{*}{\textbf{Measure}} & \multicolumn{3}{|c|}{\textbf{Inferred Parameters}} \\
\cline{2-4}
{} & {$D_{1}$} & {$\alpha$} & {$\beta$} \\
\hline
{Mean} & {$0.451$} & {$0.871$} & {$-0.890$} \\
\hline
{Variance} & {$6.2 \times 10^{-5}$} & {$2.5 \times 10^{-4}$} & {$1.5 \times 10^{-4}$} \\
\hline
{Error (\%)} & {12.7} & {3.2} & {2.2} \\
\hline
{Data Loss} & \multicolumn{3}{|c|}{$2.0$} \\
\hline
\end{tabular}
\label{tab: set_D_R}
\end{center}
\end{table}

\section{Conclusion}

The results from the different parameter sets illustrate the capability of the inverse PINN in inferring RD parameters with low data losses, thus providing both original and potential alternative parameter sets. These alternative parameters produced similar stripe periodicity, thickness and magnitude as the original parameter set when provided to our numerical model, showing that they indeed represent viable solutions to the provided desired pattern. 

Critically, our proof-of-concept results show that the PINN is able to work well across patterns of varying modes and types. Among the five parameters used for inference, $D_{1}, \alpha$ and $\beta$ were observed to have a heavier impact on pattern appearance while the system's periodicity was less sensitive to $r_{1}$. On the other hand, the results suggest that $r_{1}$ has a scaling effect on the final concentrations of $u$ and $v$, although the pattern characteristics remain similar. Ultimately, we conclude that the inverse PINN is a suitable candidate for data-driven discovery of PDE parameters in RD systems. In particular, this can be of interest to design of self-assembly systems via the guided choice of constituent compounds and reactions. Critically, the stochastic nature of the optimization also provides multiple viable parameter sets for each pattern at no additional computational cost, which can be a welcome feature for inverse design.

The proof-of-concept work here focused on steady-state inference of the parameter sets, which is particularly relevant for design. However, as observed in the results above, there can be huge variance in the underlying parameter sets if the model only has access to the steady-state pattern. Hence, we anticipate that extending this inverse PINN methodology to the modelling of RD systems with transient information may be of particular interest to the characterization and study of known biological systems where 1 specific parameter set needs to be identified, and would be interesting future work.

\section*{Acknowledgment}

J.K. was supported under the CFAR Internship Award for Research Excellence 2022. This work was also supported by A*STAR under the AI3 HTPO seed grant (Award No. C211118016) on Upside-Down Multi-Objective Bayesian Optimization for Few-Shot Design.

\bibliography{mybibfile}

\begin{thebibliography}{10}
\expandafter\ifx\csname url\endcsname\relax
  \def\url#1{\texttt{#1}}\fi
\expandafter\ifx\csname urlprefix\endcsname\relax\def\urlprefix{URL }\fi
\expandafter\ifx\csname href\endcsname\relax
  \def\href#1#2{#2} \def\path#1{#1}\fi

\bibitem{turing1990chemical}
A.~M. Turing, The chemical basis of morphogenesis, Bulletin of mathematical
  biology 52~(1) (1990) 153--197.

\bibitem{kondo1995}
S.~Kondo, R.~Asai, A reaction--diffusion wave on the skin of the marine
  angelfish {P}omacanthus, Nature 376~(6543) (1995) 765--768.

\bibitem{kondo2010}
S.~Kondo, T.~Miura, Reaction-diffusion model as a framework for understanding
  biological pattern formation, Science 329~(5999) (2010) 1616--1620.

\bibitem{asai1999}
R.~Asai, E.~Taguchi, Y.~Kume, M.~Saito, S.~Kondo, Zebrafish {L}eopard gene as a
  component of the putative reaction-diffusion system, Mechanisms of
  Development 89~(1--2) (1999) 87--92.

\bibitem{bard1981}
J.~B. Bard, A model for generating aspects of zebra and other mammalian coat
  patterns, Journal of Theoretical Biology 93~(2) (1981) 363--385.

\bibitem{yin2004}
Y.~D. Yin, R.~M. Rioux, C.~K. Erdonmez, S.~Hughes, G.~A. Somorjai, A.~P.
  Alivisatos, Formation of hollow nanocrystals through the nanoscale
  {K}irkendall effect, Science 304~(5671) (2004) 711--714.

\bibitem{boekhoven2015}
J.~Boekhoven, W.~E. Hendriksen, G.~J.~M. Koper, R.~Eelkema, J.~H. Van~Esch,
  Transient assembly of active materials fueled by a chemical reaction, Science
  349~(6252) (2015) 1075--1079.

\bibitem{mann2009}
S.~Mann, Self-assembly and transformation of hybrid nano-objects and
  nanostructures under equilibrium and non-equilibrium conditions, Nature
  Materials 8 (2009) 781--792.

\bibitem{miyazawa2010}
S.~Miyazawa, M.~Okamoto, S.~Kondo, Blending of animal colour patterns by
  hybridization, Nature Communications 1~(66) (2010) 1--6.

\bibitem{maini1997}
P.~K. Maini, K.~J. Painter, H.~N.~P. Chau, Spatial pattern formation in
  chemical and biological systems, Journal of the Chemical Society, Faraday
  Transactions 93 (1997) 3601--3610.

\bibitem{ni2005turing}
W.-M. Ni, M.~Tang, Turing patterns in the lengyel-epstein system for the cima
  reaction, Transactions of the American Mathematical Society 357~(10) (2005)
  3953--3969.

\bibitem{kaltenbacher2020inverse}
B.~Kaltenbacher, W.~Rundell, The inverse problem of reconstructing
  reaction--diffusion systems, Inverse Problems 36~(6) (2020) 065011.

\bibitem{karniadakis2021physics}
G.~E. Karniadakis, I.~G. Kevrekidis, L.~Lu, P.~Perdikaris, S.~Wang, L.~Yang,
  Physics-informed machine learning, Nature Reviews Physics 3~(6) (2021)
  422--440.

\bibitem{chiu2022can}
P.-H. Chiu, J.~C. Wong, C.~Ooi, M.~H. Dao, Y.-S. Ong, {CAN}-{PINN}: {A} fast
  physics-informed neural network based on coupled-automatic--numerical
  differentiation method, Computer Methods in Applied Mechanics and Engineering
  395 (2022) 114909.

\bibitem{raissi2019}
M.~Raissi, P.~Perdikaris, G.~E. Karniadakis, Physics-informed neural networks:
  A deep learning framework for solving forward and inverse problems involving
  nonlinear partial differential equations, Journal of Computational Physics
  378 (2019) 686--707.

\bibitem{arthurs2020}
C.~Arthurs, A.~P. King, Active training of physics-informed neural networks to
  aggregate and interpolate parametric solutions to the {N}avier-{S}tokes
  equations, arXiv:2005.05092v2 [physics.comp-ph] (2020).

\bibitem{eivazi2021}
H.~Eivazi, M.~Tahani, P.~Schlatter, R.~Vinuesa, Physics-informed neural
  networks for solving {R}eynolds-averaged {N}avier–{S}tokes equations,
  arXiv:2107.10711v1 [physics.flu-dyn] (2021).

\bibitem{giampaolo2022physics}
F.~Giampaolo, M.~De~Rosa, P.~Qi, S.~Izzo, S.~Cuomo, Physics-informed neural
  networks approach for 1{D} and 2{D} {G}ray-{S}cott systems, Advanced Modeling
  and Simulation in Engineering Sciences 9~(1) (2022) 1--17.

\bibitem{wang2021eigenvector}
S.~Wang, H.~Wang, P.~Perdikaris, On the eigenvector bias of fourier feature
  networks: {F}rom regression to solving multi-scale pdes with physics-informed
  neural networks, Computer Methods in Applied Mechanics and Engineering 384
  (2021) 113938.

\bibitem{krishnapriyan2021characterizing}
A.~Krishnapriyan, A.~Gholami, S.~Zhe, R.~Kirby, M.~W. Mahoney, Characterizing
  possible failure modes in physics-informed neural networks, Advances in
  Neural Information Processing Systems 34 (2021) 26548--26560.

\bibitem{barrio1999}
R.~Barrio, C.~Varea, J.~Arag{\'o}n, P.~Maini, A two-dimensional numerical study
  of spatial pattern formation in interacting turing systems, Bulletin of
  Mathematical Biology 61~(3) (1999) 483--505.

\bibitem{ermentrout1991}
B.~Ermentrout, Stripes or spots? nonlinear effects in bifurcation of
  reaction—diffusion equations on the square, Proceedings of The Royal
  Society A 434~(1891) (1991) 413--417.

\end{thebibliography}

\end{document}